\documentclass[conference]{IEEEtran}
\IEEEoverridecommandlockouts
\usepackage{cite}
\usepackage{amsmath,amssymb,amsfonts}
\usepackage{algorithmic}
\usepackage{graphicx}
\usepackage{textcomp}
\usepackage{xcolor}
\def\BibTeX{{\rm B\kern-.05em{\sc i\kern-.025em b}\kern-.08em
    T\kern-.1667em\lower.7ex\hbox{E}\kern-.125emX}}

\begin{document}

\title{On the Importance of Neural Membrane Potential Leakage for LIDAR-based Robot Obstacle Avoidance using Spiking Neural Networks \\

\thanks{A. Safa supervised the project as Principal Investigator and contributed to the writing of the manuscript; Z. Ali contributed to the technical developments and to the writing of the manuscript; L. Al-Amir contributed to the technical developments.}
}

\author{\IEEEauthorblockN{Zainab Ali$^1$, Lujayn Al-Amir$^1$, Ali Safa$^1$}
\IEEEauthorblockA{$^1$\textit{College of Science and Engineering, Hamad Bin Khalifa University, Doha Qatar} \\
zaal88832@hbku.edu.qa, lual88833@hbku.edu.qa, asafa@hbku.edu.qa}
}

\maketitle

\begin{abstract}
Using neuromorphic computing for robotics applications has gained much attention in recent year due to the remarkable ability of Spiking Neural Networks (SNNs) for high-precision yet low memory and compute complexity inference when implemented in neuromorphic hardware. This ability makes SNNs well-suited for autonomous robot applications (such as in drones and rovers) where battery resources and payload are typically limited. Within this context, this paper studies the use of SNNs for performing direct robot navigation and obstacle avoidance from LIDAR data. A custom robot platform equipped with a LIDAR is set up for collecting a labeled dataset of LIDAR sensing data together with the human-operated robot control commands used for obstacle avoidance. Crucially, this paper provides what is, to the best of our knowledge, a first focused study about the importance of neuron membrane leakage on the SNN precision when processing LIDAR data for obstacle avoidance. It is shown that by carefully tuning the membrane potential leakage constant of the spiking Leaky Integrate-and-Fire (LIF) neurons used within our SNN, it is possible to achieve on-par robot control precision compared to the use of a non-spiking Convolutional Neural Network (CNN). Finally, the LIDAR dataset collected during this work is released as open-source with the hope of benefiting future research.   
\end{abstract}


\section*{Supplementary Material}
Our dataset is released at: https://tinyurl.com/47pm8rfw

\section{Introduction}

The use of neuromorphic computing for robotics has gained huge interest in recent year \cite{enbodied, rev, robosnn}, due to the fact that compute power can often be limited on small robots, with limited battery power and payload capacity \cite{navion}. Hence, neuromorphic computing techniques such as Spiking Neural Networks (SNNs) are increasingly studied within robotics applications, due to their limited compute and memory complexity compared to conventional Deep Neural Networks (DNNs) when implemented in ultra-low-power neuromorphic hardware \cite{freya}. In contrast to continuous-valued neurons (such as e.g., ReLU) used in DNNs, SNNs make use of biologically-plausible \textit{spiking} neurons (such as the \textit{Leaky Integrate and Fire} neuron) \cite{lifneur} that emit binary action potentials (i.e., zeros and ones), more closely mimicking the way neurons operate in biological brains \cite{bottom}. This spiking nature of SNNs enables them to consume very low power when implemented in neuromorphic hardware since energy is consumed only when a spike is emitted \cite{odin}. Furthermore, their spiking nature removes the need for compute-expensive multiply-accumulate (MAC) operations and replaces MACs with only additions, further reducing hardware power and area consumption \cite{addonly}.
\begin{figure}[t]
    \centering
    \includegraphics[width=0.95\linewidth]{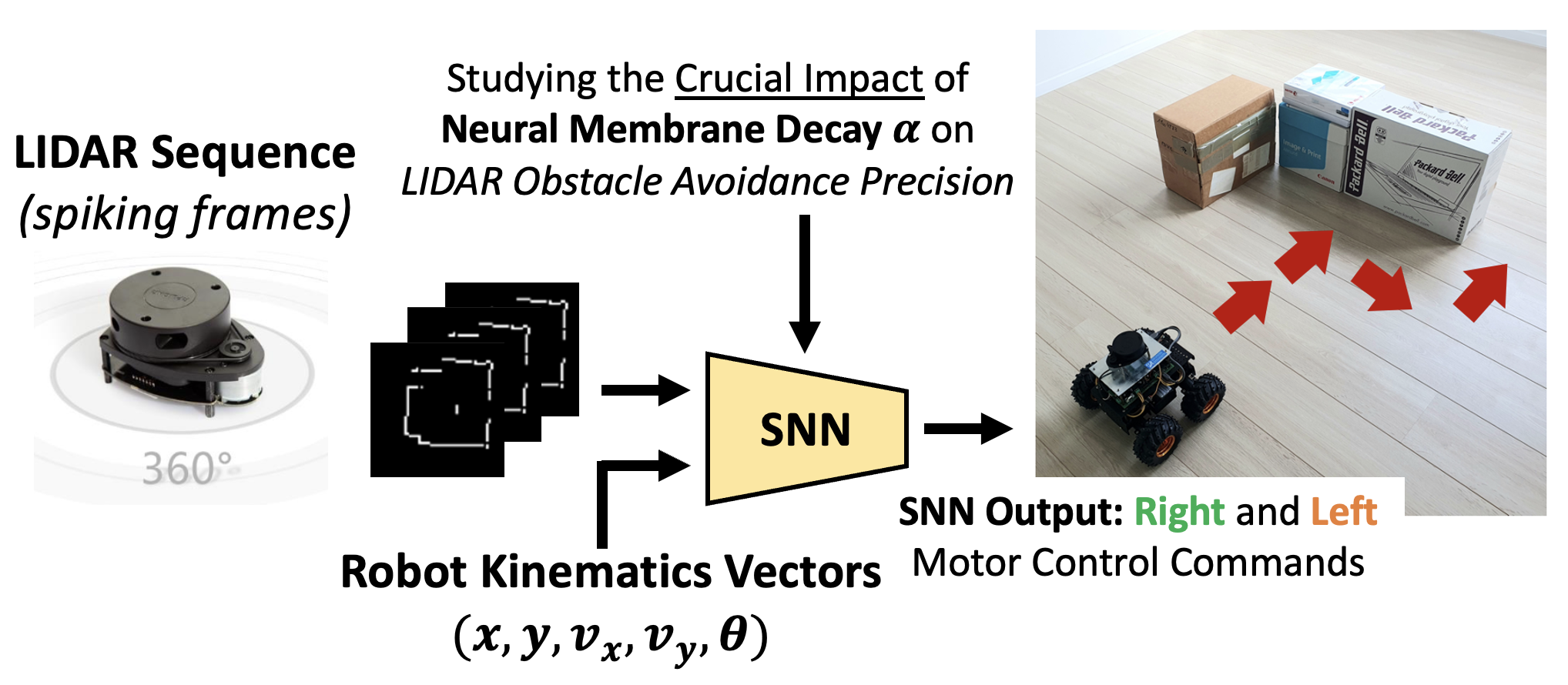}
    \caption{\textit{\textbf{Schematic view of the proposed LIDAR-SNN robot obstacle avoidance system.} In this work, we demonstrate the crucial importance of the neural membrane potential decay $\alpha$, used in the Leaky Integrate-and-Fire (LIF) spiking neurons of the SNN, on the robot obstacle avoidance precision. }}
    \label{fig:systemview}
\end{figure}
\setcounter{figure}{2}
\begin{figure*}[htbp]
    \centering
    \includegraphics[width=0.8\linewidth]{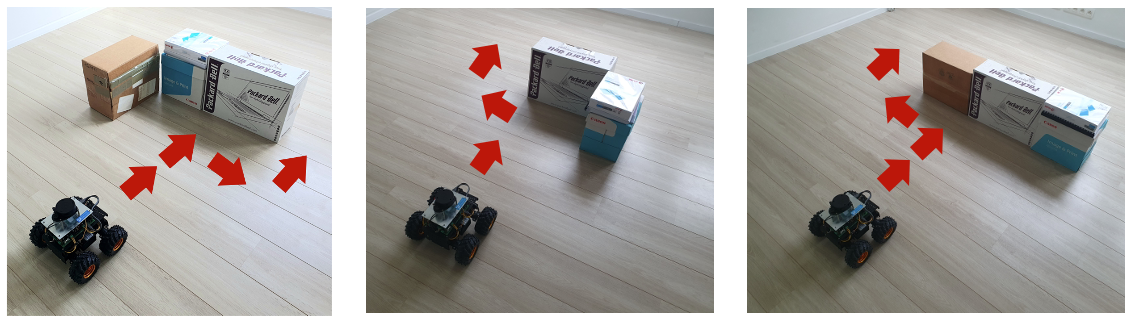}
    \caption{\textit{\textbf{Examples of obstacle avoidance paths} that the robot follows in order to reach the back side of the wall obstacles placed in the environment (see red arrows). LIDAR data collection is done by manually controlling the robot while recording the LIDAR data. The robot is manually controlled by a human operator so as to reach the back side of the obstacles in the shortest way possible. Hence the navigation path that the human operator imposes to the robot depends on the placement and configuration of the wall obstacles.}}
    \label{fig:paths}
\end{figure*}
Following these aforementioned observations on the usefulness of SNNs for robotics, this paper studies the use of SNNs for autonomous robot obstacle avoidance by processing LIDAR data as input. The use of SNNs for LIDAR processing has recently gained attention, with a number of emerging work focusing mostly on data classification \cite{lidarclassification}, object detection \cite{lidarobjectrec} and autonomous lane keeping \cite{lane,lane2}. More closely related to our work, the authors in \cite{control} propose a neuromorphic motor control system that uses LIDAR data within a \textit{Proportional-Integrate-Derivative} (PID)-like regulation system \textit{emulated} via an SNNs for collision avoidance. But to the best of our knowledge, the use of \textit{end-to-end} SNN learning for deriving motor commands directly from LIDAR and kinematics data for \textit{indoor robot obstacle avoidance} has not received significant attention yet. 

In contrast to these aforementioned works, we propose a convolutional SNN architecture that fuses LIDAR data as well as kinematics information vectors (i.e., position and velocity estimates) returned by the robot in order to infer the commands that need to be sent to the robot's right and left motor channels (see Fig. \ref{fig:systemview}) in an indoor setting. Crucially, we shed light to the critical importance of the dynamical properties of the spiking neurons used within the SNN. In particular, we experimentally demonstrate that the choice of \textit{neural membrane potential leakage constants} \cite{leakage} has a significant effect on robot navigation accuracy when processing LIDAR data with SNNs. The contributions of this paper are:
\begin{enumerate}
    \item Using a custom robot setup equipped with a LIDAR sensor, we collect a labeled dataset of more than 38 acquisition sequences of LIDAR and kinematics data, together with the remote commands sent by the human operator to the robot during data acquisition.
    \item We study the use of a convolutional SNN architecture for performing autonomous robot obstacle avoidance and we provide what is, to the best of our knowledge, a first demonstration of the crucial importance of neural membrane potential leakage constants on the LIDAR processing accuracy of the SNN.  
    \item We benchmark our proposed SNN architecture to a corresponding DNN with a similar architecture and we show that by carefully tuning the leakage constant $\alpha$ of Leaky Integrate and Fire (LIF) neurons, our proposed SNN outperforms its corresponding DNN in terms of obstacle avoidance accuracy, while leading to large savings in terms of compute complexity and hardware resources.
    \item We release our novel dataset as open source to help future research.
\end{enumerate}

This paper is organized as follows. Our robot platform and data collection setup is described in Section \ref{robotplatf}. Our SNN-based LIDAR processing method is covered in Section \ref{methodss}. Experimental results are shown in Section \ref{expresult}, and conclusions are provided in Section \ref{conc}.

\section{Robot platform and data collection}
\label{robotplatf}

This Section presents the custom robot setup that is used to collect LIDAR data for the path planning and collision avoidance task studied in this work. 
\subsection{Custom robot platform}
\label{customplatform}
Fig. \ref{fig:systemview} shows our rover robot setup that is used to record LIDAR data in order to obtain a \textit{labeled} dataset composed of \textit{a)} LIDAR frames, \textit{b)} robot internal kinematic states (position and velocity), and \textit{c)} motor control commands that will be used as target labels during the SNN training. Hence, the SNN will process the LIDAR data and internal robot state vectors as input in order to infer the motor commands that will drive the robot around the obstacles, while following the shortest path depending on the obstacle configuration. 
\setcounter{figure}{1}
\begin{figure}[htbp]
    \centering
    \includegraphics[width=1\linewidth]{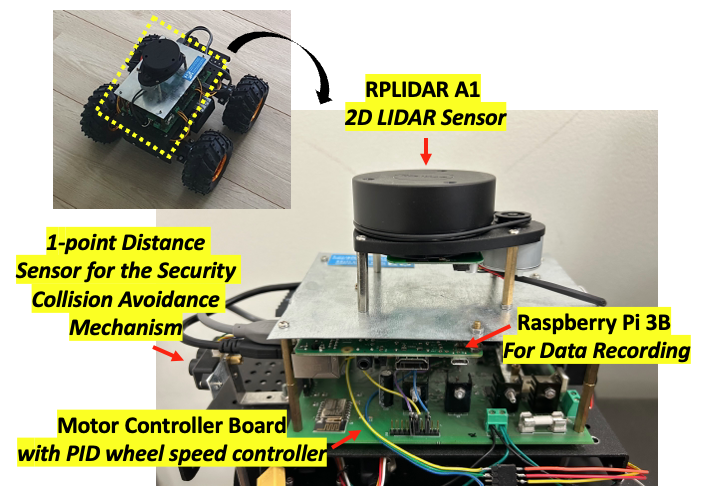}
    \caption{\textit{\textbf{Custom robot system used for LIDAR data collection.} The robot is equipped with an RPLIDAR A1 2-dimensional LIDAR sensor connected to a Raspberry Pi board that is used to record the LIDAR frames. The robot motors are controlled via a custom robot controller board previously presented in \cite{robocontrol} which provides motor speed control using a Proportional-Integral-Derivative (PID) controller and security stopping via a 1-point distance sensor (to prevent the risk of collisions). The robot is connected to the WIFI network and is remotely controlled via a laptop through the WIFI connection.}}
    \label{fig:systemview}
\end{figure}
\setcounter{figure}{3}
The rover setup in Fig. \ref{fig:systemview} features an \textit{RPLIDAR A1} providing 2-dimensional LIDAR detection point clouds. The data is recorded using a \textit{Raspberry Pi} that is connected to both the LIDAR and the motor controller board. The motor controller board  is a custom-made PCB containing the various electronics circuits required for driving the motors while precisely controlling their rotation speed via an on-board Proportional-Integral-Derivative (PID) controller \cite{robocontrol}. The motor controller board also returns wheel odometry data (i.e., the measured wheel rotation speeds) to the Raspberry Pi, and is also connected to a front-facing 1-point distance sensor for securely shutting down the motors in case a collision is about to happen. 

\subsection{LIDAR data collection for obstacle avoidance}
\label{datasetdesc}

We use the rover setup presented in Section \ref{customplatform} to collect a labeled dataset for the study of LIDAR-based robot obstacle avoidance with SNNs. Our dataset collection is done as follows. First, the rover is placed in front of a number of \textit{wall obstacles} (consisting of cardboard boxes). The configuration of the wall obstacles can be changed by arbitrary placing the cardboard boxes in different positions on the floor. Then, a human operator remotely drives the robot with the goal of reaching the back side of the obstacles while following the shortest path towards this goal position. At the same time, the LIDAR data $\{\mathcal{L}_k\}$ is recorded during the obstacle avoidance process, together with the remote driving commands $\{ \mathcal{C}_k \}$ that were sent by the human operator (where $k$ denotes the time step during data recording). 

The driving commands $\mathcal{C}_k$ consist of \textit{2-dimensional} vectors where each entry denotes the \textit{rotation direction} (positive or negative) of the right and left motor channels. As such, the driving command vectors $\mathcal{C}_k$ can only take four different values: $(1,1), (-1,1), (1,-1)$ and $(-1, -1)$. In all cases, the absolute motor rotation \textit{speed} is set to $2\pi$ rad/s.

In addition, the internal robot kinematic state vectors $\{\mathcal{S}_k\}$ containing the robot position $x,y$, velocity $v_x,v_y$ and pose angle $\theta$ estimates are also recorded during the data collection. The robot position and velocity are estimated using an \textit{Extended Kalman Filter} (EKF) that fuses the \textit{wheel odometry} data with \textit{LIDAR odometry} estimates obtained using a conventional \textit{Iterative Closest Point} (ICP) algorithm.

Doing so, a labeled data sequence $\{\mathcal{L}_k,\mathcal{S}_k,\mathcal{C}_k\}$ is obtained for each data acquisition run, where the remote driving commands $\mathcal{Y}_k=\mathcal{C}_k$ act as the target labels that need to be estimated by the SNN given the LIDAR and robot state input $\mathcal{X}_k = \{\mathcal{L}_k,\mathcal{S}_k\}$.

During our data acquisition campaign (see Fig. \ref{fig:paths}), we collect a total of $38$ obstacle avoidance sequences, with each acquisition sequence having a duration of about $\sim 1$ minute each. The recording frame rate for the LIDAR data $\{\mathcal{L}_k\}$, kinematics $\{\mathcal{S}_k\}$ and label commands $\{\mathcal{C}_k\}$ is set to $10$ fps.

The acquisitions forming the dataset are recorded using six different wall obstacle configurations consisting of straight walls and angled walls with different wall width sizes and placement configurations (see Fig. \ref{fig:paths}). This makes the collected dataset diverse in terms of wall obstacle placement. Furthermore, the different obstacle configurations are recorded in a way so as to make the dataset \textit{balanced} in terms of acquired scenarios (i.e., the same amount of acquisitions were conducted for each obstacle scenario under test). Finally, our collected dataset is also released as open source as part of the supplementary materials.

\begin{figure}[t]
    \centering
    \includegraphics[width=0.7\linewidth]{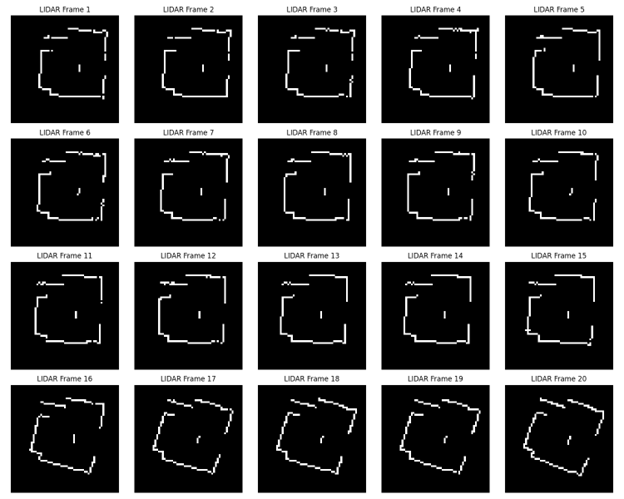}
    \caption{\textit{\textbf{Sequence of LIDAR Spike Frames Captured During Robot Navigation.} Each frame represents a snapshot of the robot's perception of its surroundings as it navigates through the environment. The white cells indicate spike-based representation of obstacles while the black cells indicate absence of spikes.} }
    \label{fig:examp}
\end{figure}

\begin{figure*}[htbp]
    \centering
    \includegraphics[width=0.95\linewidth]{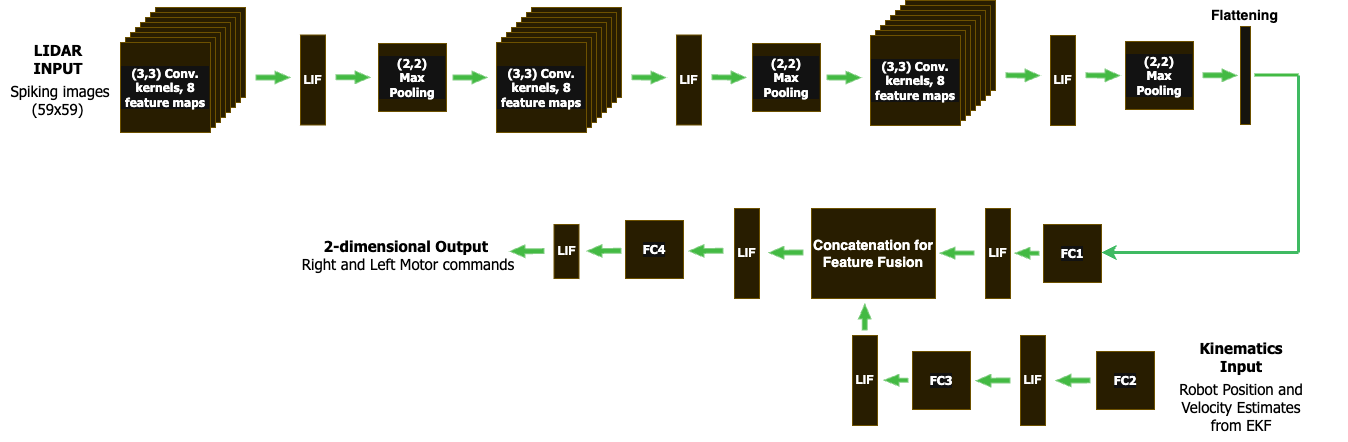}
    \caption{\textit{\textbf{SNN Architecture block diagram.} The SNN takes as input the spiking LIDAR images and the kinematics vector (robot position and velocity). The spiking LIDAR images are processed by three convolutional spiking layers with LIF neurons followed by $(2\times2)$ max pooling. The resulting spiking feature map is flatten as a 1-dimensional vector and processed by a fully-connected (FC1) weight matrix, before being concatenated with the feature vector at the output of FC3 that is obtained by processing the kinematics data. This concatenated fusion vector is then processed by a final spiking layer (FC4) and a 2-dimensional spiking vector is obtained as output, driving the rotation direction of the right and left motors in the robot. }}
    \label{fig:SNN_Architecture_used}
\end{figure*}

\section{Methods}
\label{methodss}

This Section first describes the pre-processing method used to convert the input LIDAR data into a spiking representation. Then, an overview of the SNN architecture used in this work together with its training approach is provided.

\subsection{LIDAR data pre-processing}

As the data obtained from RPLIDAR A1 sensor is in a continuous-valued format, it needs to be converted into a format of binary spike-based inputs that mimic neuron firing so that the data can be made compatible with the spiking nature of SNNs. Raw LIDAR data consists of continuous values such as distance and angle measurements. Each detection $d$ returned by the LIDAR consists of polar coordinates \((r_d, \theta_d)\) where \(r_d\) is the distance from the sensor to the object and \(\theta_d\) is the angle of the detected object relative to the sensor. These polar coordinates are then converted into Cartesian coordinates \((x_d, y_d)\) where \(x_d = r_d \times cos(\theta_d)\) and \(y_d = r_d \times sin(\theta_d)\). 

After coordinate transformation, LIDAR data points are mapped on a 2-dimensional image plane to create a grid-like representation. The grid acts like a pixel map where each pixel represents either 0 or 1. The value of 0 indicates the absence of obstacle while the value of 1 indicates an obstacle. The grid size is $59 \times 59$ columns where each cell represents a small spatial region. This results in the conversion of the raw LIDAR data sequence into a \textit{sequence of spiking images}. 

As the LIDAR collects data over time, each scan generates a new binary spike image. In our dataset, we have recorded 38 navigation sessions with each session containing around $\sim 200$ frames, thus ensuring balance and consistency during model training and batch processing. Since the robot continuously captures its surroundings as it moves, each frame captures a snapshot of obstacle positions at that moment. These frames are stacked to form a sequence which allows the SNN to capture time-domain dynamic changes in the obstacle positions due to the movement of the robot over time. Fig. \ref{fig:examp} shows an example of the obtained spiking LIDAR image sequence from the first 20 frames of the first data sequence in the dataset. 

\subsection{SNN architecture}
 
Our SNN architecture in Fig. \ref{fig:SNN_Architecture_used} takes as input both spiking LIDAR images $\{\mathcal{L}_k\}$ representing the spatial layout and the obstacles in the robot's vicinity, as well as the robot's internal kinematic state vectors $\{\mathcal{S}_k\}$ capturing the robot's position and velocity (see dataset description in Section \ref{datasetdesc}). The SNN outputs 2-dimensional motor control command vectors $\Bar{s}_{out} = [r, l]$ capturing the rotation directions $r,l \in \{1,-1\}$ of the right and left motor channels. It is important to note that since the SNN uses spiking neurons, its output will take $0$'s and $1$'s as possible values, but we remap an output value of $0$ to $-1$ in order to represent a negative motor rotation direction as $-1$ during the experiments.

As spiking neuron model, we make use of the popular \textit{Leaky Integrate-and-Fire} (LIF) neuron due to its tradeoff between computational simplicity while modeling well spiking neural dynamics compared to more simplistic models such as \textit{Integrate-and-Fire} (I\&F). Upon receiving its input current $J=\Bar{w}^T \Bar{s}_{in}$ resulting from the inner product between the neuron's weight vector $\Bar{w}$ and the vector of input spikes $\Bar{s}_{in}$, the LIF neuron integrates its input current $J$ through the time-step iterations $k$ as:
\begin{equation}
    V_{k+1} \xleftarrow{} \alpha V_k + (1-\alpha) J
\end{equation}
where $V$ is the membrane potential and $\alpha$ is the membrane potential \textit{decay constant} governing the leakage rate of $V$. Then, as soon as $V$ crosses a certain threshold $\theta$ (set to $\theta=1$ in this work), the neuron emits an output spike $s_{out}=1$ and the membrane potential is reset back to zero.


\subsubsection{Architectural choices}
The architectural choices made during the design of the SNN model are motivated as follows. First, we use convolutional layers to process the input LIDAR data in order to better capture spatial patterns in the spiking LIDAR images. We also make use of Max Pooling layers which keep the spiking nature of the data after the pooling operation \cite{snntrain}. Pooling is performed at the output of the spiking neurons in each of the three convolution layers. Indeed, if Average Pooling would be used, the output values after pooling would be continuous (between $0$ and $1$). However, Max Pooling ensures that the output values are \textit{either} 0 or 1, making it compatible with neuromorphic hardware implementation.

In addition to the spiking LIDAR image processing path, the SNN includes a second input path for processing the robot's kinematic state vectors. At each time step \(k\), the kinematics input vector \(\mathcal{S}_k \in {\mathbb{R}}^5\) is first processed through a fully-connected layer (FC2) which feeds into a layer of spiking LIF neurons. Doing so, the kinematics vectors are converted into a spiking representation at the output of the FC2-LIF layer in Fig. \ref{fig:SNN_Architecture_used}. Then, this spiking representation is processed by another fully-connected layer FC3 followed by LIF neurons which produces a 32-dimensional spiking feature vector out of the robot kinematics state vectors. At the stage, the spiking feature vector representing the kinematics data is concatenated with the flattened spiking embedding obtained from the LIDAR data though the convolutional layers. Hence, this concatenated \textit{fusion} vector encodes both the robot's perception and its internal motion information in the spiking domain.  

Finally, the 2-dimensional output spiking vector $\Bar{s}_{out}$ representing the commands that need to be sent to the robot's motors is derived by processing the fusion spiking vector through the fully-connected layer FC4 and a pair of output LIF neurons. The output vector $\Bar{s}_{out} \in \{0,1\}$ is remapped as follows:
\begin{equation}
    \Bar{s}_{out} \xleftarrow{} (2\Bar{s}_{out} - 1)
\end{equation}
which remaps the output values from $\{0,1\}$ to $\{-1,1\}$, representing the negative and positive motor rotation directions.

\subsubsection{Training approach}

We make use of the popular \textit{surrogate gradient} method for training our SNN, which replaces the ill-defined gradient $\sigma'(V)$ of the output of a spiking neuron in function of its membrane potential by a Gaussian surrogate function:
\begin{equation}
    \sigma'(V)\approx \frac{1}{\sqrt{2\pi}}e^{-2V^2}
    \label{surrogate}
\end{equation}

Since the SNN model includes time dependicies due to the state of the LIF neurons, training is carried out using the Back-Propagation-Through-Time (BPTT) algorithm. In practice, we instantiate our SNN architecture using the \textit{pytorch} library and by defining custom neuron models that behave as the LIF non-linearity during inference but make use of the surrogate gradient (\ref{surrogate}) during the backpropagation phase.

As loss function, we use the \textit{Mean Squared Error} (MSE) to quantify the difference between predicted and target output values. 
\begin{equation}
    L_{\mathrm{MSE}}=\frac{1}{N_{batch}} \sum_{i=1}^{N_{batch}}\left(\mathcal{Y}_{i}-\Bar{s}_{out,i}\right)^{2}
\end{equation}
where \(\mathcal{Y}_{i}\) are the label commands and \(\Bar{s}_{out,i}\) are the predicted output values from the model. 

\section{Results}
\label{expresult}
In this section, we report our experimental results and compare the performance of our SNN architecture to its corresponding CNN model. In all our experiments, we use 5-fold cross-validation with a $80\%$-$20\%$ train-test data split in order to report both the average precision (reported as the MSE loss evaluated on the test set) and its standard deviation. The SNN is trained over $90$ epochs using a batch size of $16$ using the Adam optimizer (with its default parameters) and with a learning rate of $\eta=10^{-3}$.

\subsection{Impact of the LIF membrane potential leakage}

\begin{figure}[htbp]
    \centering
    \includegraphics[width=1\linewidth]{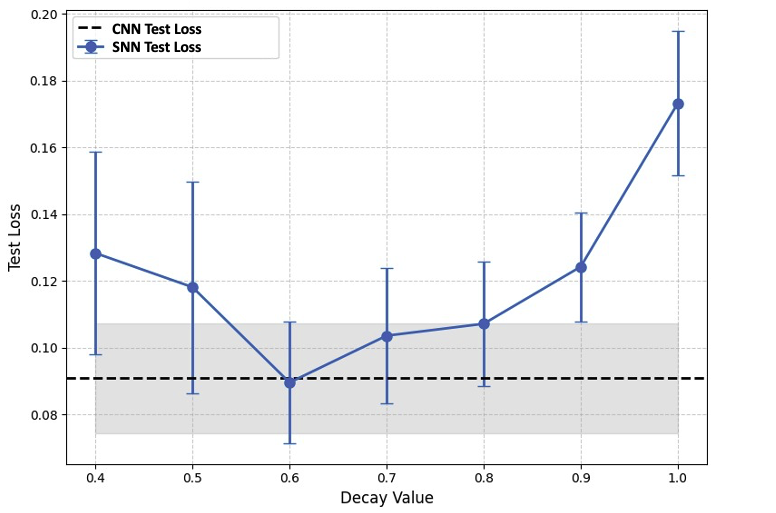}
    \caption{\textit{\textbf{Precision of SNN vs. CNN (test loss).} The average SNN test loss and its standard deviation is reported in function of the LIF neuron membrane decay value $\alpha$ ($80\%$-$20\%$ train-test split with $5$-fold cross-validation). For reference, the average CNN accuracy and its standard deviation are also shown on the graph. The best SNN test loss is obtained for $\alpha=0.6$. The p-value statistics between the SNN and the CNN is $p=0.9646 \geq 0.05$, denoting that there is no statistically significant differences between the SNN and CNN test precisions.}}
    \label{fig:res2}
\end{figure}

Fig. \ref{fig:res2} reports the SNN precision (test loss) in function of the LIF membrane decay $\alpha$. It can be seen that the membrane decay value is optimal near $\alpha=0.6$, leading to a minimum average SNN test loss of $l_{test}^{snn}=9 \times 10^{-2}$ with a standard deviation of $\sigma_{test}^{snn} = 3.5 \times 10^{-2}$. This clearly demonstrates the crucial importance of LIF membrane decay value when processing LIDAR data with our SNN, within our robot obstacle avoidance task. Indeed, if e.g., an \textit{Integrate-and-Fire} (I\&F) neuron would be used ($\alpha=1$ in Fig. \ref{fig:res2}), this would \textit{degrade} the SNN precision by more than $\times 2$ larger test loss.

\begin{figure}[htbp]
    \centering
    \includegraphics[width=1\linewidth]{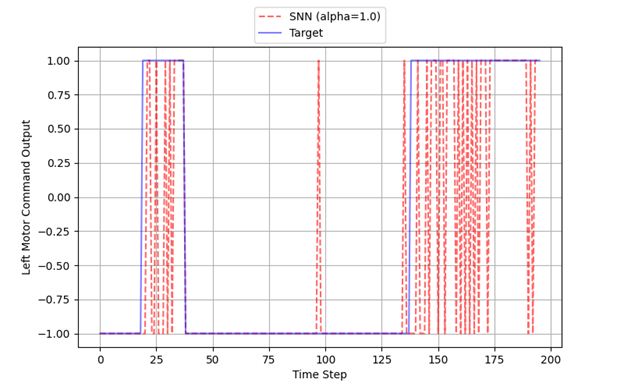}
    \caption{\textit{\textbf{SNN output example with $\alpha=1$ (no leakage).} The SNN output is affected by many spurious spikes, degrading the model precision.} }
    \label{fig:enter-label2}
\end{figure}

\begin{figure}[htbp]
    \centering
    \includegraphics[width=1\linewidth]{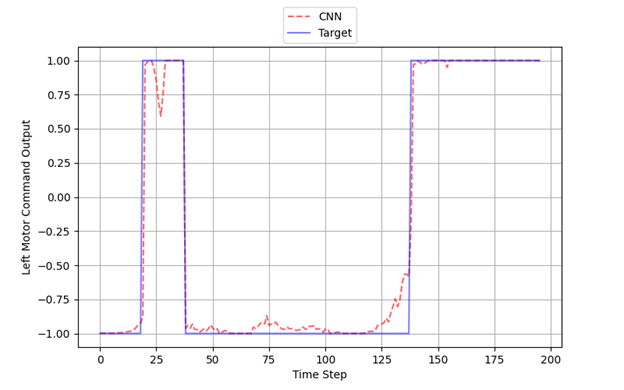}
    \caption{\textit{\textbf{CNN output (left motor command channel) vs. target label command.} The CNN output can take continuous values between $-1$ and $1$, while the SNN output values are always restricted to $-1$ or $1$ due to its spiking nature.}}
    \label{fig:cnnoutputstyle}
\end{figure}

\begin{figure}[htbp]
    \centering
    \includegraphics[width=1\linewidth]{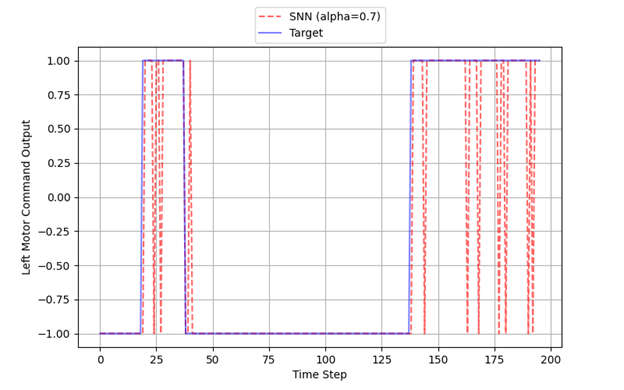}
    \caption{\textit{\textbf{SNN output example with $\alpha=0.6$ (no leakage).} Compared to the case of Fig. \ref{fig:enter-label2} ($\alpha=1$), a significantly smaller amount of spurious spikes affect the SNN output, leading to better model precision. } }
    \label{fig:enter-label3}
\end{figure}

The importance of the membrane decay can be explained as follows. We observed during our experiments that due to the spiking nature of the SNN output (vs. \textit{continuous} nature of the CNN output), an error in the SNN output always boils down to a spurious (or outlier) output spike (see Fig. \ref{fig:enter-label2}). These spurious spikes have a strong impact on the calculated test loss since they induce a large deviation compared to an error in the CNN output, since it can take any values between $-1$ and $1$ (see Fig. \ref{fig:cnnoutputstyle}) instead of being restricted to $-1$ \textit{or} $1$ as in the SNN case. Within this context, reducing the LIF membrane decay constant $\alpha$ makes the neurons within the SNN \textit{more leaky} and less prone to generating a large amount of spurious output spikes (see Fig. \ref{fig:enter-label3}). On the other hand, if $\alpha$ is kept on being reduced, the leakage within the SNN becomes too strong which delays the output and degrades the overall SNN performance.

\subsection{SNN vs. CNN performance via Welch's t-test}
\label{snnvscnn}
Fig. \ref{fig:res2} also reports the average CNN test loss and its standard deviation, which are respectively $l_{test}^{cnn}=9.1 \times 10^{-2}$ and $\sigma_{test}^{cnn} = 3.4 \times 10^{-2}$. In order to analyze how statistically significant is the difference in precision between the SNN (at $\alpha=0.6$) and the CNN, we resort to the standard \textit{Welch's t-test}, given $\mu_1=9 \times 10^{-2}$ (average SNN test loss), $\sigma_1=3.5 \times 10^{-2}$ (SNN standard deviation), $\mu_2=9.1 \times 10^{-2}$ (average CNN test loss), $\sigma_2=3.4 \times 10^{-2}$ (CNN standard deviation) and $n_1=n_2=5$ (the number of cross-validation runs used to compute the average precisions). 

By applying Welch's t-test, we obtain a \textit{t-value} of $t=0.046$ and a \textit{p-value} of $p=0.9646$. Since the obtained p-value is larger than $0.05$, it can be concluded that there is \textit{no statistically significant differences} between the SNN and CNN test precision. This shows that when using $\alpha=0.6$, our SNN setup performs \textit{on par} with its equivalent CNN architecture, \textit{while being implementable on neuromorphic hardware}.

\subsection{SNN vs. CNN compute complexity}

\begin{figure}[htbp]
    \centering
    \includegraphics[width=0.9\linewidth]{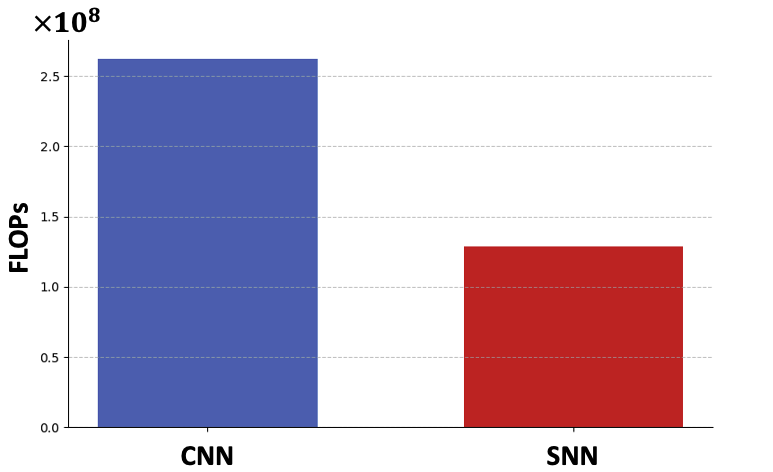}
    \caption{\textit{\textbf{Number of Floating Point Operations (FLOPs) per inference.}  } }
    \label{fig:comp}
\end{figure}

We evaluate the reduction in compute complexity, in terms of the number of floating point operations (FLOPs), that the SNN enables when implemented in neuromorphic hardware. Indeed, thanks to its use of spiking neurons (instead of continuous-valued neurons in CNNs), it is well known that the matrix multiplications (i.e., \textit{multiply-accumulate} operations) required at the level of each weight layer reduces to simple \textit{additions only} \cite{noadd}, reducing the overall FLOPs (see Fig. \ref{fig:comp}). 

Fig. \ref{fig:comp} shows that using our proposed SNN significantly reduces the model compute complexity compared to its CNN counterpart while \textit{not sacrificing} the model precision (see Section \ref{snnvscnn}). Even though the analysis in Fig. \ref{fig:comp} only considered the FLOP reduction that would be attainable when implementing our proposed SNN in neuromorphic hardware, it is important to mention that further reduction in both energy consumption and chip area utilization compared to CNNs could be obtained using e.g., emerging sub-threshold mixed-signal SNN accelerator hardware \cite{mixedsig,dynapse}, closely mimicking the event-driven nature of information processing in the brain. 

\subsection{Comparison with prior works}

Finally, it is important to position our results relative to prior research. Among the most recent related work is by A. Shalumov \textit{et al.} \cite{control}, who proposed a neuromorphic PID controller using an SNN to steer a LIDAR-equipped car in a simulated outdoor environment (vs. real-world indoor data in our work). While direct comparison is difficult due to differing contexts (i.e., outdoor car navigation in \cite{control} vs. indoor robot obstacle avoidance in our work), our approach fundamentally differs: we use an end-to-end SNN that directly maps LIDAR data to robot commands. In contrast, \cite{control} employs a multi-step, PID-like SNN scheme that neither explores direct LIDAR-to-control learning nor analyzes spiking neuron parameters. Hence, the end-to-end nature of our SNN significantly reduces system complexity, enabling single-SNN deployment on neuromorphic hardware, \textit{unlike} the heterogeneous multi-step system in \cite{control}.

Additionally, R. Zhu \textit{et al.} \cite{lane2} proposed a deep SNN for outdoor autonomous driving and lane keeping, based on a spiking ResNet. However, their model targets outdoor car navigation and is unsuitable for our energy-constrained indoor robot due to its high resource demands. Moreover, \cite{lane2} does not assess the role of LIF neuron parameters in LIDAR processing, which is a central focus of our work.

\section{Conclusion}
\label{conc}
This paper has investigated the use of SNNs for autonomous robot navigation and obstacle avoidance using LIDAR and kinematics data, which were collected using a custom robot setup equipped with a 2D LIDAR sensor along with control commands issued by a human operator. Our proposed SNN model has been designed to predict motor commands based on LIDAR spiking images and kinematics state vector. In particular, the key contribution of this paper lies in the analysis of the membrane potential leakage constant \(\alpha\) in LIF neurons. It has been shown that carefully tuning the membrane potential leakage constant within our SNN has a crucial impact on its inference precision, as it helps reducing spurious spikes and leading to an enhanced control accuracy. In addition, it has been shown that the proposed SNN architecture can attain on-par precision compared to its equivalent CNN by choosing a well-suited leakage constant $\alpha$, while being significantly less compute-expensive. Finally, the dataset used in this work has been released with the hope of benefiting future research.

\end{document}